\documentclass[review]{elsarticle}
\pdfoutput=1
\usepackage{enumitem}

\usepackage{algorithm}
\usepackage{algorithmic}
\usepackage{setspace}

\usepackage{float}
\usepackage{multirow}
\usepackage{amssymb}
\usepackage{makecell}
\usepackage{threeparttable}
\usepackage{booktabs}
\usepackage{threeparttable}
\usepackage{lineno,hyperref}

\journal{Information Sciences}

\begin{document}

\begin{frontmatter}

\title{Proximal Policy Optimization via Enhanced Exploration Efficiency}


\author[mymainaddress,mysecondaryaddress]{Junwei Zhang}

\author[mymainaddress,mysecondaryaddress]{Zhenghao Zhang}
\author[mymainaddress,mysecondaryaddress]{Shuai Han}

\author[mymainaddress,mysecondaryaddress]{Shuai L\"u\corref{mycorrespondingauthor}}
\cortext[mycorrespondingauthor]{Corresponding author}
\ead{lus@jlu.edu.cn}

\address[mymainaddress]{Key Laboratory of Symbolic Computation and Knowledge Engineering (Jilin University), Ministry of Education, Changchun 130012, China}
\address[mysecondaryaddress]{College of Computer Science and Technology, Jilin University, Changchun 130012, China}

\begin{abstract}
	
Proximal policy optimization (PPO) algorithm is a deep reinforcement learning algorithm with outstanding performance, especially in continuous control tasks. But the performance of this method is still affected by its exploration ability. For classical reinforcement learning, there are some schemes that make exploration more full and balanced with data exploitation, but they can't be applied in complex environments due to the complexity of algorithm. Based on continuous control tasks with dense reward, this paper analyzes the assumption of the original Gaussian action exploration mechanism in PPO algorithm, and clarifies the influence of exploration ability on performance. Afterward, aiming at the problem of exploration, an exploration enhancement mechanism based on uncertainty estimation is designed in this paper. Then, we apply exploration enhancement theory to PPO algorithm and propose the proximal policy optimization algorithm with intrinsic exploration module (IEM-PPO) which can be used in complex environments. In the experimental parts, we evaluate our method on multiple tasks of MuJoCo physical simulator, and compare IEM-PPO algorithm with curiosity driven exploration algorithm (ICM-PPO) and original algorithm (PPO). The experimental results demonstrate that IEM-PPO algorithm needs longer training time, but performs better in terms of sample efficiency and cumulative reward, and has stability and robustness.
\end{abstract}


\end{frontmatter}

\section{Introduction}

Deep reinforcement learning combines the efficient representation and perception ability in deep learning with the decision-making ability in reinforcement learning, which can learn and optimize the policies via interacting with complex environments. This approach, which is closer to the human mode of thinking, has achieved remarkable results in arcade games \cite{HesselMHSODHPAS18,HasseltGS16}, physical simulation environments \cite{HaarnojaZAL18,WuMGLB17} and real tasks \cite{MahmoodKVMB18,isci/LiYXCR20}, but is also facing with more complex modeling and training processes.

In the application of reinforcement learning algorithm, the environment data should be processed according to actual situation. According to action characteristics, simulation environment tasks can be divided into discrete action tasks and continuous action tasks. Although both two action types can be generated by corresponding neural networks \cite{GuLSL16,IqbalS19}, the outputs and the performance will be greatly different due to different model architectures.

In continuous action tasks, proximal policy optimization algorithm \cite{SchulmanWDRK17} proposed by Schulman in 2017 performs well. It has the advantages of stable training process, high performance and scalability. However, some researchers also find it difficult to reproduce PPO algorithm, which mainly due to the sensitivity of hyperparameters, and agent could not balance the exploration of the environment and the exploitation of data \cite{abs-1811-02553}, so that the agent's policy easily falls into local optima.

The Gaussian action exploration mechanism in PPO samples actions from a Gaussian distribution, which adds a certain range of variance as the noise to the output of policy network. Such an exploration mechanism samples actions with equal probability in opposite directions from the current mean, which can stably optimize the policy towards the optimal decision direction theoretically. However, due to the limited number of interactions with the environment and the estimation error \cite{CiosekVLH19}, this mechanism tends to fall into the local optima. We will illustrate the relationship between the exploration scope and the suboptimal solution by experiments with different exploration settings.

In physical simulation environment, we construct deep neural network to estimate the uncertainty of action. And the uncertain value of observation state is used to stimulate action selection, so as to encourage the action to explore in the direction of great uncertainty and avoid converging prematurely to local optima. Our approach also provides a trend for agent to explore further. This mechanism is applied to PPO algorithm, and we call it IEM-PPO (proximal policy optimization with intrinsic exploration module).

In the experiment, we use MuJoCo tasks, which have been widely recognized in recent years. While reproducing PPO algorithm, we optimize and adjust the algorithm settings according to the experimental environment. The expected results are achieved in all tasks, and in some tasks are better than those given in PPO literature \cite{SchulmanWDRK17}. We apply curiosity driven mechanism \cite{PathakAED17} to PPO algorithm for the comparative experiment. The experiment demonstrates that although the exploration enhancement algorithm proposed in this paper requires longer training time, it performs better in  terms of sample efficiency and cumulative reward, and has stronger stability and robustness.

Our main contributions are as follows: 1) We analyze the defect of Gaussian noise exploration and show that exploration enhancement mechanism can play a certain role in dense reward continuous tasks. 2) Based on the uncertainty theory, an effective mechanism of exploration enhancement is proposed. On Mujoco benchmarks, our algorithm outperforms previous algorithms and more stable.

\section{Related Work}

At present, most practical deep reinforcement learning methods still rely on simple exploration rules, such as $\epsilon$-greedy in DQN \cite{MnihKSRVBGRFOPB15}, Gaussian noise in DDPG  \cite{LillicrapHPHETS15} and PPO, and entropy regularization in A3C \cite{MnihBMGLHSK16} to prevent premature convergence in the early stage of training. This kind of methods hardly introduce deviation in the optimization process, but low efficiency, especially in continuous action tasks, easy to fall into the local optima.

For exploration and exploitation, Pathak et al. introduce curiosity driven mechanism into Actor-Critic framework of deep reinforcement learning in view of the most complex environments in Atari \cite{PathakAED17}. They define curiosity as the error in the perception of environmental changes, and combine such error as internal incentive with external reward of environment as the training data. Their algorithm performs well in the discrete tasks with sparse reward \cite{BurdaEPSDE19}. However, it is difficult to predict environmental change which is influenced by bias and noise. The performance is limited when curiosity driven algorithm is applied to continuous tasks with dense reward, such as MuJoCo. Performance comparison can be seen in our experiment section.

In reinforcement learning with table method, it has been found that the uncertainty of observation state is negatively correlated with the number of states visited. Therefore, Auer et al. \cite{AuerCF02} and Stehl et al. \cite{StrehlL08} respectively propose the theory of exploration enhancement based on uncertainty in 2002 and 2008, which could theoretically fully explore the environment and maintain consistency of the optimal policy. However, after the combination of deep learning and reinforcement learning, some exploration mechanism cannot be used in complex tasks due to the time and space complexity.

In addition, in DQN, noise can be added to the parameter space of the neural network to assist the exploration \cite{FortunatoAPMHOG18}. However, in PPO and other policy gradient algorithms, the current policy is directly optimized without maintaining global Q function, so it's difficult to apply the noise in the parameter space.

\section{Background}

Reinforcement learning, as one of the important branches of machine learning, is mainly aimed at decision-making and optimization problems. Without accurate label data, the optimal policies can be obtained by trial and error through interaction with the environment.

\subsection{Problem description} 

In order to solve reinforcement learning problem, it is usually modeled as Markov decision process (MDP). MDP is a mathematical model to deal with sequential decision making problem, which is applied to simulate the stochastic policies and rewards of an agent in an environment with Markov property. MDP is usually abstracted as a tuple representation $< S,A,P,R,\gamma >$, where:

\begin{itemize}
	\item $S$ is a set of all states in the environment;
	\item $A$ is a set of executable actions of the agent;
	\item $P$ is a transition distribution, $P_{ss^{'}}^{a} = P\left( S_{t + 1} = s^{'} \middle| S_{t} = s,A_{t} = a \right)$;
	\item $R$ is a return function, and $R_{t}$ represents the environmental return obtained after taking an action at time $t$;
	\item $\gamma$ is a discount factor, which is used to calculate the cumulative reward, where $\gamma \in \left\lbrack 0,1 \right\rbrack$.
\end{itemize}

Agent selects an action to execute by action selection policy which we need to optimize. The policy is a probability distribution of actions under given state, represented by $\pi\left( a \middle| s \right)$, where $\pi\left( a \middle| s \right) = P\left( A_{t} = a \middle| S_{t} = s \right)$. That is the probability of agent chooses action $a$ under the current state $s$ at time $t$.

The goal of reinforcement learning is to maximize the reward value, which is usually calculated by using cumulative reward function $G_{t}$, defined as formula (\ref{100}).
\begin{equation}
G_{t} = R_{t} + {\gamma R}_{t + 1} + \gamma^{2}R_{t + 2} + \ldots = {\sum\limits_{k = 0}^{\infty}{\gamma^{k}R_{t + k}}}\label{100}
\end{equation}
where, the discount factor $\gamma$ is added to adjust the weighting between immediate and delayed outcomes. The greater discount factor is, the more focus agent puts on delayed outcomes. Cumulative reward function $G_{t}$  has become one of the criteria to evaluate policy in MDP problems \cite{WuMGLB17,SchulmanWDRK17,MnihBMGLHSK16}.
	
\subsection{Policy gradient} 

The policy gradient as a way to find optimal policy, samples environment interaction of agent and calculates the gradient of current policy directly, then optimizes the current stochastic policy \cite{SuttonMSM99}.

In the policy gradient algorithm, the process from the starting to the termination of the task is called an episode $\tau$, where $\tau = \left\{ s_{1},a_{1},s_{2},a_{2},\ldots,s_{T},a_{T} \right\}$. We need to maximize the cumulative reward function for all episodes. Therefore, the optimal policy is shown in formula (\ref{2}).
\begin{equation}
\pi^{*} = {argmax}_{\pi}E_{\tau\sim\pi(\tau)}\left\lbrack R\left( \tau \right) \right\rbrack\label{2}
\end{equation}

Therefore, the policy value is shown in formula (\ref{3}).
\begin{equation}
L\left( \theta \right) = E_{\tau\sim\pi_{\theta}(\tau)}\left\lbrack {R\left( \tau \right)} \right\rbrack = {\sum_{\tau\sim\pi_{\theta}(\tau)}{P_{\theta}\left( \tau \right)R\left( \tau \right)}}\label{3}
\end{equation}
where, $\theta$ is the current model parameter, $P_{\theta}\left( \tau \right) = {\prod\limits_{t = 1}^{T}{\pi_{\theta}\left( a_{t} \middle| s_{t} \right)P\left( a_{t + 1} \middle| s_{t},a_{t} \right)}}$ is occurrence probability of current trajectory. Since environment transition distribution is independent of model parameters, $P\left( a_{t + 1} \middle| {s_{t},a_{t}} \right)$  is not considered. After calculating the gradient of the objective function $L\left( \theta \right)$, formula  (\ref{4}) is shown as following:
\begin{equation}
\nabla_{\theta}L\left( \theta \right) \approx {\sum_{\tau\sim\pi_{\theta}{(\tau)}}{R\left( \tau \right)\nabla_{\theta} {log}\pi_{\theta}\left( \tau \right)}} \approx \frac{1}{N}{\sum\limits_{n = 1}^{N}{\sum\limits_{t = 1}^{T}{R\left( \tau^{n} \right)\nabla_{\theta}{log}\pi_{\theta}\left( a_{t}^{n} \middle| s_{t}^{n} \right)}}}\label{4}
\end{equation}

Therefore, policy gradient algorithm is divided into following two steps for continuous iterative update:
\begin{enumerate}[fullwidth,itemindent=2em,label=\arabic*)]
	\item Use $\pi_{\theta}$ to interact with environment, and obtain the observed data and calculate $\nabla_{\theta}L\left( \theta \right)$.
	\item Update $\theta$ with gradient and learning rate $\alpha$, where $\tilde{\theta} = \theta + \alpha\nabla_{\theta}L\left( \theta \right)$. 
\end{enumerate}

\subsection{Deep learning and variance control} 

In the complex environment, due to time and space complexity of algorithm, it is no longer possible to maintain the value table of each state-action pair. Therefore, it is necessary that agent utilizes powerful representation ability of neural network to calculate the value function and make decision \cite{XuZH14}.

In the simple environment, an accurate estimation of policy gradient can be achieved with a small number $N$ of sampled trajectories. Due to the increase of environment complexity, the demand for the number of sample trajectories also increase. In order to ensure the performance in tasks, $N$ is hard to get enough. In order to solve the problem of variance increase due to insufficient sampling \cite{SchulmanMLJA15}, the strategy of variance reduction is usually used to ensure the accuracy of gradient estimation. This variance reduction strategy is based on such two ideas: 

\begin{enumerate}[fullwidth,itemindent=2em,label=\arabic*)]
	\item The action at current moment $t$  has nothing to do with reward at the past moment $t^{'}$. At $t > t^{'}$, there is $E\left\lbrack \nabla_{\theta}\pi_{\theta}\left( a_{t} \middle| s_{t} \right)R_{t}^{'} \right\rbrack = 0$.
	\item  Adding or subtracting constant $V(s_{t})$ to current state value $R_{t}$  will not introduce bias into policy gradient, and variance can be reduced, that is $E\left\lbrack \nabla_{\theta}\pi_{\theta}\left( a_{t} \middle| s_{t} \right)V( s_{t} ) \right\rbrack = 0$. Therefore, ${\hat{A}}_{t} = {\sum\limits_{t^{'} = t}^{T}\left( R_{t^{'}} - V( s_{t} ) \right)}$  is used to replace $R\left( \tau^{n} \right)$  in the calculation of policy gradient.
\end{enumerate}

Thus, the more accurate optimization objective function is shown in formula (\ref{5}).
\begin{equation}
\nabla_{\theta}L\left( \theta \right) = \frac{1}{N}{\sum\limits_{n = 1}^{N}{\sum\limits_{t = 1}^{T}{{\hat{A}}_{t}^{n}\nabla_{\theta}{log}\pi_{\theta}\left( a_{t}^{n} \middle| s_{t}^{n} \right)}}}\label{5}
\end{equation}
where, ${\hat{A}}_{t}$ is called advantage value, representing the difference between the value of action taken and the average value in the current state. In training process, average value of current state cannot be calculated, but can be fitted by using supervised learning based on deep neural network. Therefore, two neural network approximators are needed to calculate the policy gradient in training. One is to compute $\pi_{\theta}$, which is used to select actions, the other is to compute $V( s_{t} )$ , which gives average value of current state and is used to calculate ${\hat{A}}_{t}$.

\subsection{Proximal policy optimization algorithm} 

The policy gradient algorithm is implemented by calculating estimated value of policy gradient and using stochastic gradient ascent optimization. Formula (\ref{6}) is taken as the form of policy estimation:
\begin{equation}
\nabla_{\theta}L\left( \theta \right) = E_{t}\left\lbrack {\nabla_{\theta}{\mathit{log}{\pi_{\theta}\left( a_{t} \middle| s_{t} \right)}}{\hat{A}}_{t}} \right\rbrack\label{6}
\end{equation}
where, $\pi_{\theta}$ is stochastic policy, ${\hat{A}}_{t}$ is estimation of advantage value. Expected value $E_{t}$ is calculated by averaging the set of sampling values. The algorithm continuously alternates in sampling and optimization. This estimation $\nabla_{\theta}L\left( \theta \right)$ is the derivative of objective function (\ref{7}).
\begin{equation}
L^{PG}\left( \theta \right) = E_{t}\left\lbrack {{log{\pi_{\theta}\left( a_{t} \middle| s_{t} \right)}}{\hat{A}}_{t}} \right\rbrack\label{7}
\end{equation}

In order to improve training efficiency and make sampled data reusable \cite{SchulmanWDRK17}, $\pi_{\theta}\left( a_{t} \middle| s_{t} \right)/\pi_{\theta_{old}}\left( a_{t} \middle| s_{t} \right)$ is used instead of $\mathit{\log}{\pi_{\theta}\left( a_{t} \middle| s_{t} \right)}$ to support off-policy training. However, in this way, the difference between new and old policies should not be too large. Otherwise, many samples are needed to get the correct  estimation due to the increase of variance.

It can be seen from objective function that: when ${\hat{A}}_{t} > 0$ , the policy will optimize in the direction that $\mathit{\log}{\pi_{\theta}\left( a_{t} \middle| s_{t} \right)}$ increases, thence the probability of selecting the current action $a_{t}$ will increase. The size of training step is determined by the combination of action selection probability $\pi$, advantage function $\hat{A}$ and learning rate $\alpha$. If the learning rate is limited to a small range to ensure policy changes little, the probability of policy falling into local optima will increase.

In PPO algorithm, a clipping mechanism is added to objective function to punish the policy change, when $\pi_{\theta}\left( a_{t} \middle| s_{t} \right)/\pi_{\theta_{old}}\left( a_{t} \middle| s_{t} \right)$ is away from 1, which represents policy updating generated by the maximized objective function is excessive. Final objective function is shown in formula (\ref{8}).
\begin{equation}
L^{CLIP}\left( \theta \right) = {\hat{E}}_{t}\left\lbrack {min\left( r_{t}\left( \theta \right){\hat{A}}_{t},clip\left( \epsilon,r_{t}\left( \theta \right) \right){\hat{A}}_{t} \right)} \right\rbrack\label{8}
\end{equation}
where, $r_{t}\left( \theta \right) = \pi_{\theta}\left( a_{t} \middle| s_{t} \right)/\pi_{\theta_{old}}\left( a_{t} \middle| s_{t} \right)$ represents action selection probability ratio of new and old policies. In the $min$ operation, the first item is the original optimization goal, and the second item is $clip$ function which modifies the surrogate objective by clipping the probability ratio. The $clip$ function removes the incentive for moving  $r_{t}$ outside of the interval $\left\lbrack 1 - \epsilon,1 + \epsilon \right\rbrack$, where $\epsilon$ is hyperparameter. In this scheme, only large changes in the direction of policy improvement are removed, while in the policy decline are retained. According to the comparing results of the Mujoco experiments \cite{SchulmanWDRK17}, $\epsilon$ is set to 0.2. The performance is less affected by $\epsilon$ values. The PPO algorithm is shown in Algorithm \ref{alg:A}.

\begin{algorithm}
	\caption{PPO}
	\label{alg:A}
	\begin{algorithmic}[1]
		\STATE {Initial policy parameters $\theta_{0}$ and value   function parameters $\phi_{0}$. } 
		\FOR{$k$ = 0,1,2, …} 
		\STATE{Collect set of trajectories $\mathcal{D}_{k} = \left\{ \tau_{i} \right\}$ by running policy $\pi_{k} = \pi\left( \theta_{k} \right)$ in the environment.}
		\STATE{Compute   reward-to-go  ${\hat{R}}_{t}$.}
		\STATE{Compute   advantage estimation ${\hat{A}}_{t}$ based on value function $V_{\phi}$.}
		\STATE{Update the policy with Adam by maximizing the clip objective:
		\begin{center}
		$\theta_{k + 1} = {{\arg\max\limits_{\theta}}{\frac{1}{\left| \mathcal{D}_{k} \right|T}{\sum\limits_{ \tau \in \mathcal{D}_{k}}{\sum\limits_{t = 0}^{T}{min}}}\left( {\frac{\pi_{\theta}\left( a_{t} \middle| s_{t} \right)}{\pi_{\theta_{k}}\left( a_{t} \middle| s_{t} \right)}{\hat{A}}_{t},clip\left( \epsilon,r_{t}\left( \theta \right) \right){\hat{A}}_{t}} \right)}}.$
		\end{center}}
		\STATE{Approximate   value function via gradient descent  algorithm by regression on mean-squared error:   
		\begin{center}
		$\phi_{k + 1} = {{\arg\min\limits_{\phi}}{\frac{1}{\left| \mathcal{D}_{k} \right|T}{\sum\limits_{ \tau \in \mathcal{D}_{k}}{\sum\limits_{t = 0}^{T}\left( {{\hat{R}}_{t}- V_{\phi}\left( s_{t} \right) } \right)^{2}}}}}.$
		\end{center}}
		\ENDFOR
	
	\end{algorithmic}
\end{algorithm}

\section{Enhance exploration ability}

In discrete action tasks with sparse reward, the relative insufficiency of exploration ability can be obviously detected. However, for continuous action tasks with dense reward, the problem of balancing between exploration and exploitation still exists. Original exploration mechanism may not enough to make exploration and exploitation reach a balanced situation.

\subsection{Exploration and exploitation of continuous action tasks} 

In reinforcement learning environments, policy optimization needs to maximize cumulative reward. For any state $s$, depending on the number of state visits, the degree of perception of alternative actions is usually different. Even when the policy reward reach convergence, there are also unselected or less selected actions, and the evaluation of such actions is often inaccurate. In training process, exploitation is to use the current policy in order to obtain better reward value with current perception; exploration is trying different actions to gather more information. A good long-term training usually needs sacrifice short-term gains by gathering enough information to enable an agent to get a global optimal solution. Although exploration and exploitation are contradictory processes, a good exploration strategy can improve performance and an efficient exploitation strategy can also increase the scope of exploration.

When using PPO algorithm to solve continuous action tasks, the policy network selects an appropriate action $\mu_{t}$ according to the acquired knowledge. In order to explore unknown actions, Gaussian noise $\mathcal{N}\left( 0,\sigma \right)$ is added to action $\mu_{t}$ to obtain $a_{t}$, as shown in formula (\ref{9}). The $a_{t}$ is used as the final action to interact with environment.
\begin{equation}
a_{t} = \mu_{t} + \mathcal{N}\left( 0,\sigma \right)\label{9}
\end{equation}
Therefore, this kind of action selection with noise is classified as a stochastic policy in continuous action space. According to mathematical characteristics of Gaussian distribution, formula (\ref{10}) can be used to obtain likelihood probability of taking current action $a_{t}$:
\begin{equation}
{log{P\left( a_{t} \right)}} = {\sum\limits_{i = 1}^{n}{- \frac{1}{2}\left( {\left( \frac{a_{ti} - \mu_{ti}}{{\sigma}_{i}} \right)^{2} + 2{\mathit{log}{\sigma}_{i}} + {log}2\pi} \right)}}\label{10}
\end{equation}

The value of $\sigma$ in Gaussian noise will determine the ability to select unknown actions in current policy. The larger $\sigma$ tends to select unknown actions, which will lead to problems of low training efficiency and policy instability, while stable policy get higher scores in most tasks. If a smaller $\sigma$ value is chosen, exploration ability is poor, the policy would fall into local optima. Therefore, $\sigma$ should be selected as hyperparameter of a certain problem.

According to optimization theory: In initial stage of neural network approximating optimization, giving a large learning rate of parameters, the algorithm can have a stronger ability to jump out of local optima. It tends to choose a more appropriate optimization point in global. As the learning process continues to reduce learning rate, the ability to approximate optimal value becomes stronger. PPO algorithm could use a dynamic $\sigma$ setting, cumulative reward is used as the index to measure learning progress. Exploration scope decreases with the increase of cumulative reward, which has a better training effect. Since cumulative reward values of the task within a fixed range, training effects are still different for different initial values. If inappropriate exploration parameters are given, the whole experiment process will be affected.

\begin{figure}[ht]
	
	\centering
	\includegraphics[scale=0.25]{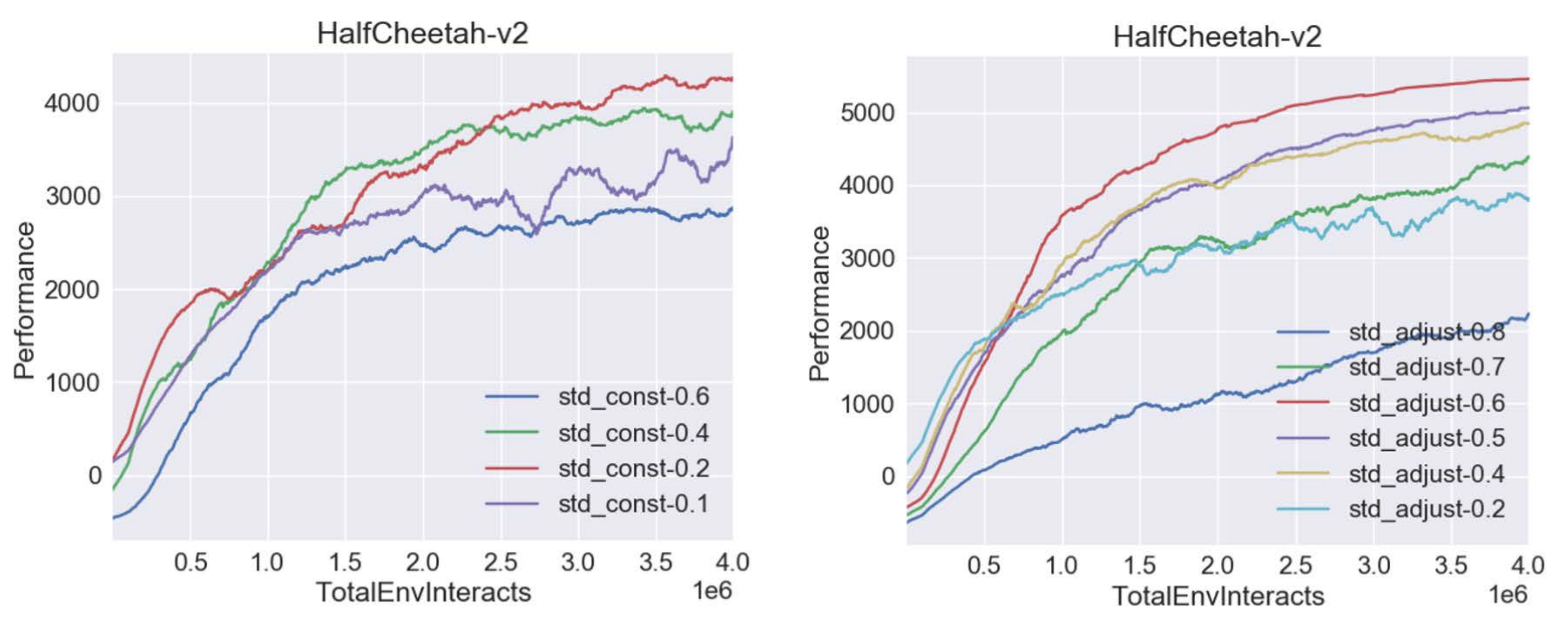}       
	\caption{Experiments of various fixed $\sigma$ settings (left) and dynamic $\sigma$ value settings (right) in Halfcheetah-v2 environment}
	\label{fig1}
	
\end{figure}

We use experiments to verify the theory. The experimental environment is Halfcheetah-v2 in MuJoCo. Details of the environment will be introduced in the experimental section. The experimental results of fixed $\sigma$ and dynamic $\sigma$  settings based on cumulative reward are shown in Figure \ref{fig1}.

In Figure \ref{fig1}, training results of four fixed exploration value settings \{0.6, 0.4, 0.2, 0.1\} and training results of six dynamic  $\sigma$ settings under different initial values of training \{0.8, 0.7, 0.6, 0.5, 0.4, 0.2\} are respectively shown. The abscissa represents the number of interactions with the environment, the ordinate represents the sum of reward of each episode. It can be seen from the experimental results that the final performance set by dynamic  $\sigma$ can converge to 5000, which is better than the fixed $\sigma$ settings of 4000. And in different settings, the problem of falling into local optimum is common. For the Halfcheetah-v2 task, the dynamic $\sigma$ setting with the initial value of 0.6 is better, all other settings present some degree of local optimum problem. 

\subsection{Curiosity driven mechanism} 

In some tasks, for example, \textit{Montezuma's revenge} in Atari 2600, there are situations where external reward available to agent is sparse. However, for reinforcement learning tasks, an agent constantly explores environment until it gets a reward, and then can improve the policy so that it can get better reward and learn policies in future. Hoping to get a reward by chance (i.e. random exploration) is likely to be futile for all but the simplest of environments     \cite{PathakAED17}.

In curiosity theory, internal incentive signals add to the original environmental reward, and it uses existing data information to expand the original sparse reward for reasonable reward form dense, which assists agent with better exploration. Formally expressed as formula (\ref{11}).
\begin{equation}
R_{t} = R_{t}^{i} + R_{t}^{e}\label{11}
\end{equation}

The addition of internal incentive signals are intended to stimulate the direction to novel states, which increases the probability of revisiting and further exploration in this direction during policy learning. Therefore, internal incentive signals mainly determine the specification for whether a state is novel in a complex environment. At modeling, measuring ``novelty"  requires a statistical model of the distribution of environmental states, that is, getting how novel of the current state after giving certain observations.

In curiosity module, curiosity is defined as the error of predicting the result of one's own behavior, in other words, the perception of agent on the degree of the environment affected by current actions. In modeling, the forward prediction model is joined, current state and action as neural network input. We use the multilayer neural network, output next state observation values and minimize mean square error (MSE) as training target. Prediction model gets internal incentive signals and trains positive ability to predict observation state. The mixed reward is shown in formula (\ref{12}).
\begin{equation}
R_{t} = R_{t}^{i} + \beta\left\| {f_{\psi}\left( s_{t},a_{t} \right) - s_{t + 1}} \right\|_{2}^{2}\label{12}
\end{equation}
where, $f_{\psi}\left( s_{t},a_{t} \right)$ is forward neural network with parameter $\psi$, $\beta$ is hyperparameter which serves as the balance ratio between internal incentive and external reward, which needs experimental adjustments to eliminate dimensional influence.

Previously, the curiosity module is mainly aimed at tasks with insufficient external reward signals, but theoretically, these internal incentive signals give a new direction of exploration in the short term. In continuous action tasks, the introduction of curiosity module points to a new mechanism for enhancing exploration. Different from original Gaussian noise, internal incentive can encourage the exploration of novel states for a short time. If internal incentive and Gaussian random action can act together, directional exploration enhancement can be guaranteed under the condition of less action noise.

\begin{figure}[ht]
	\centering
	\includegraphics[scale=0.35]{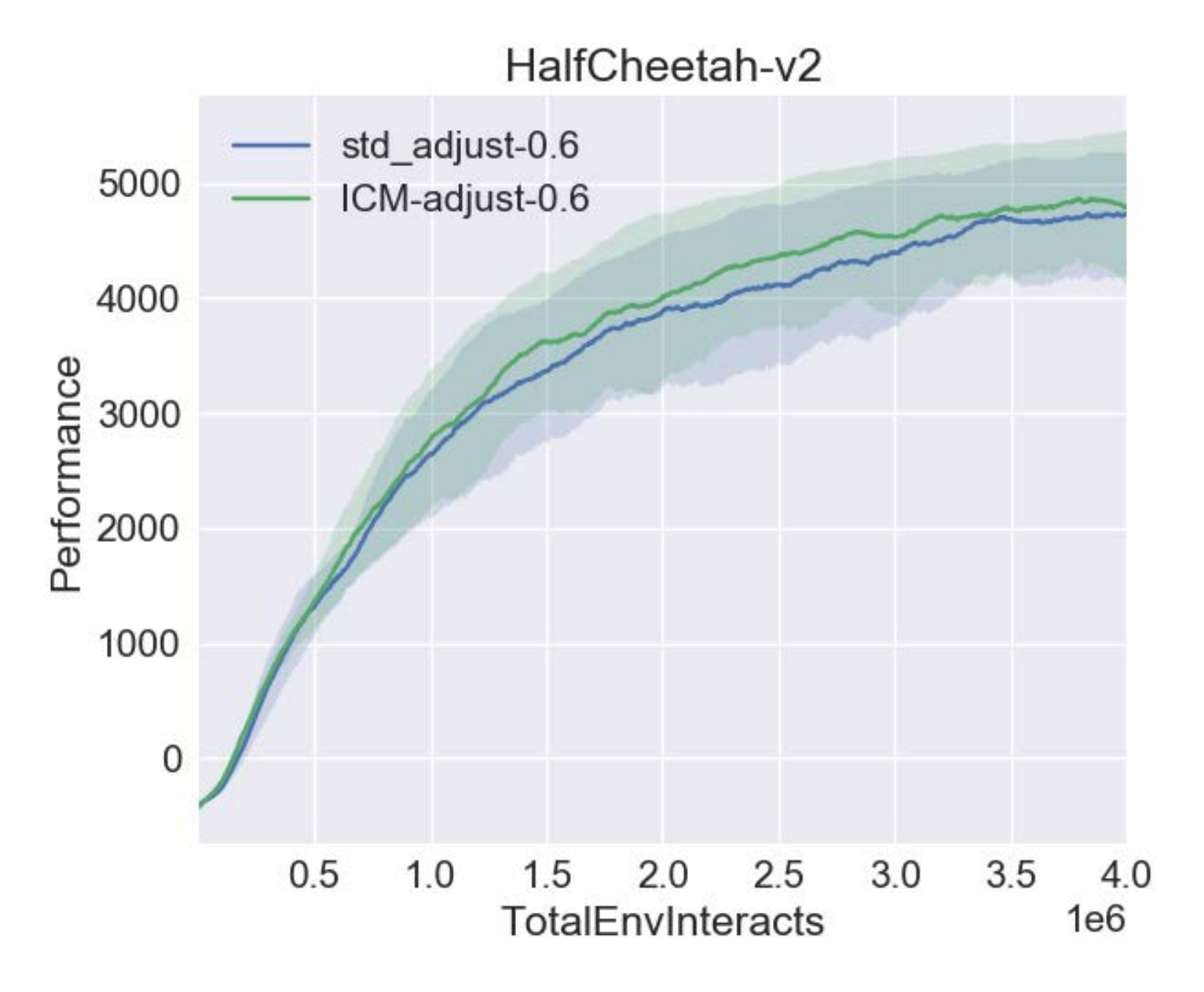}       
	\caption{Comparison experiment of curiosity driven algorithm (ICM-PPO) and PPO algorithm in Halfcheetah-v2 environment}
	\label{fig2}	
\end{figure}

In order to demonstrate the performance of curiosity module, we use Halfcheetah-v2 in MuJoCo to carry out experimental test. Experimental results are shown in Figure \ref{fig2}.

In Figure \ref{fig2}, curiosity driven algorithm and PPO algorithm run three times respectively. The results are averaged and the variance is marked with shadows. It can be seen that after the introduction of curiosity module, training speed in the early stage has little impact, but in the middle and late stages, due to the full perceptive learning of state which fewer visit in original algorithm, it achieves a higher cumulative reward. Although improvement is small, it can be explained that curiosity driven module, as a directional guide to exploration mechanism, has certain effects on continuous action tasks and can play a role in enhancing training efficiency. Curiosity algorithm is taken as one of the experimental comparison algorithms in this paper, more complete experiments and analyses will be given in the experiment section.

\subsection{Exploration enhancement based on uncertainty estimation} 

Just as previous analysis of PPO algorithm, original Gaussian noise exploration has fallen into a dilemma, which is mainly summarized as the following two points:

\begin{enumerate}[fullwidth,itemindent=2em,label=(\arabic*),listparindent=2em]
	\item Gaussian exploration is sensitive to hyperparameters.
	
	The performance is seriously affected by the initial value in both fixed and dynamically adjusted $\sigma$. Moreover, when initial values are small, it is easy to fall into the local optima; when initial values are large, the exploration scope is increased, and the policy will be unstable which ultimately is difficult to converge.

	\item  Exploration efficiency is insufficient.
	
	In the process of action selection, Gaussian noise exploration mechanism samples actions with equal probability in opposite directions from the current mean. In order to ensure exploration ability, initial exploration scope must be set as a large value. Some actions are no longer necessary for further exploration, repeating exploration slows down learning efficiency.
	
\end{enumerate}

Therefore, for continuous action tasks, a directional exploration mechanism that can be used for low-Gaussian noise action selection will improve exploration ability of the algorithm. We design an enhanced exploration algorithm by analyzing exploration mechanism of table method.

Prior to the rise of deep learning and its integration with reinforcement learning, reinforcement learning used tabular methods to record updates and iterations of state and action value. Although deep reinforcement learning performs well in various tasks, most of mechanisms in the original reinforcement learning cannot be transferred into deep reinforcement learning due to the limitations of data processing, such as uncertain behavior exploration.

Uncertain behavior exploration method relis on the concept of information entropy. Action reward is represented by distribution function. If the variance of distribution function is large, then it contains more information; if the variance of action reward distribution is small, it requires fewer visits to determine actual action reward. Therefore, in order to obtain the maximum exploration ability, action reward distribution with large variance should be explored first.

Since uncertain behavior exploration always gives priority to exploration of actions with great uncertainty, it is also called optimistic exploration. There is relatively great uncertainty in the states with few visits. It is potentially believed that there is a better policy scheme in unknown actions. The bonus of the state $s$ is added based on the number of visits $N( s )$, as shown in formula (\ref{13}).
\begin{equation}
r^{+}\left( {s,a} \right) = r( {s,a}) + B( N( s ) )\label{13}
\end{equation}

In 2008, Stehl et al. \cite{StrehlL08} consider the impact of interaction times on the uncertainty. On the basis of improving the exploration efficiency, algorithm ensures the environment with no bias of optimal policy and is more convenient to calculate, as shown in formula (\ref{14}).
\begin{equation}
B( {N( s)} ) = \sqrt{\frac{1}{N( s)}}\label{14}
\end{equation}

As the training progresses, the number of state visits $N(s )$ increases and eventually approaches infinity, so the uncertain reward gradually returns to 0, which ensures the consistency of optimal policy with the reward function $r^{+}( {s,a} )$ and $r( {s,a} )$  in theory.

For the problem of too many environmental states, deep neural network as an extension of function approximation completes the work of action selection and action value estimation in each state. For the uncertainty estimation of complex environment, the powerful representation ability of deep neural network can also be used to design the internal incentive operator $\hat{N}( s )$, which is related to the number of visits, from the data recorded statistically, to achieve the function similar to that of tabulation statistical calculation, as shown in formula (\ref{15}).
\begin{equation}
\hat{N}( s ) \propto B( {N( s )} )\label{15}
\end{equation}

Although some random bias is introduced in the estimation of deep neural network, the error can be ignored in the appropriate uncertainty estimation.

In model, for training stage, current state $s_{t}$ and subsequent state $s_{t + n}$ are combined into a vector as input. Neural network estimates value $\hat{n}$ of the required action steps, and uses real required steps $n$ as the loss function for approximating learning. For sampling stage, the number of steps that required to complete current change can be estimated by using current state $s_{t}$ and next state $s_{t + 1}$. The real number of current steps required is 1. If the number of steps required is large, it indicates that current state transition is difficult to complete, needs to be stimulated with a high degree of uncertainty to encourage agents to explore more into current state transition mode. By motivating state transition which requires more steps, the exploration mechanism also makes agent more inclined to choose actions that make environment more changeable, so as to promote agent to explore more broadly states.

The combination of uncertain reward and external reward forms the mixed reward of agent training, as shown in formula (\ref{16}).
\begin{equation}
r^{+}( {s,a} ) = r( {s,a} ) + c_{1}\hat{N}( s )\label{16}
\end{equation}
where, $c_{1}$ is a hyperparameter, which is used to balance the role ratio of uncertainty reward and external environment reward in training, since dimensions of reward are different. By combining the enhancement exploration module with PPO algorithm, PPO with intrinsic exploration module (IEM-PPO) is formed. The algorithm flow is shown in Algorithm \ref{alg:B}.

Based on uncertainty estimation, IEM-PPO gives novel actions with great environmental impact. With the proportion of exploration increasing, uncertainty reward gradually decrease. Finally, exploration of most environmental states are completed and the more nearly optimal action selection policy is found. Although neural network cannot achieve optimization policy which is completely consistent with global optimal solution, the wide exploration alleviates the problem of falling into local optima in tasks and finds a suitable action selection policy.

\begin{algorithm}[H]
	\caption{IEM-PPO}
	\label{alg:B}
	\begin{algorithmic}[1]
		\STATE {Initial policy parameters $\theta_{0}$, value   function parameters $\phi_{0}$ and uncertainty estimation function parameters $\xi_{0}$. } 
		\FOR{$k$ = 0,1,2, …} 
		\STATE{Collect set of trajectories $\mathcal{D}_{k} = \left\{ \tau_{i} \right\}$ by running policy $\pi_{k} = \pi\left( \theta_{k} \right)$ in the environment.}
		\STATE{ Compute   reward-to-go  ${\hat{R}}_{t}$ based on the current uncertainty estimation function  $N_{\xi}\left( {s_{t},s_{t + 1}} \right)$  and environmental return $r$.}
		\STATE{Compute   advantage estimation ${\hat{A}}_{t}$ based on value function $V_{\phi}$.}
		\STATE{Update the policy with Adam by maximizing the clip objective:
			\begin{center}
				$\theta_{k + 1} = {{\arg\max\limits_{\theta}}{\frac{1}{\left| \mathcal{D}_{k} \right|T}{\sum\limits_{ \tau \in \mathcal{D}_{k}}{\sum\limits_{t = 0}^{T}{min}}}\left( {\frac{\pi_{\theta}\left( a_{t} \middle| s_{t} \right)}{\pi_{\theta_{k}}\left( a_{t} \middle| s_{t} \right)}{\hat{A}}_{t},clip\left( \epsilon,r_{t}\left( \theta \right) \right){\hat{A}}_{t}} \right)}}.$
		\end{center}}
		\STATE{Approximate   value function via gradient descent  algorithm by regression on mean-squared error:   
			\begin{center}
				$\phi_{k + 1} = {{\arg\min\limits_{\phi}}{\frac{1}{\left| \mathcal{D}_{k} \right|T}{\sum\limits_{ \tau \in \mathcal{D}_{k}}{\sum\limits_{t = 0}^{T}\left( {{\hat{R}}_{t}- V_{\phi}\left( s_{t} \right) } \right)^{2}}}}}.$
		\end{center}}
	\STATE{Approximate uncertainty estimation function via gradient descent algorithm by regression on mean-squared error:
		\begin{center}
			$\xi_{k + 1} = {{\arg\min\limits_{\xi}}{\frac{1}{\left| \mathcal{D}_{k} \right|T}{\sum\limits_{ \tau \in \mathcal{D}_{k}}{\sum\limits_{t = 0}^{T}\left( {n - N_{\xi}\left( {s_{t},s_{t + n}} \right)} \right)^{2}}}}}.$
	\end{center}}
		\ENDFOR
	\end{algorithmic}
\end{algorithm}

\section{Experiments} 

We have theoretically analyzed the effect and feasibility of exploration enhancement mechanism for policy gradient algorithm. Now, experimental environment will be selected, and detailed experimental comparison will be made among PPO algorithm, curiosity driven algorithm (ICM-PPO) and intrinsic exploration module algorithm (IEM-PPO). We analyze the advantages and disadvantages of each algorithm from the practical point of view.

\subsection{Experimental setup and model building} 

\begin{figure}[ht]
	\centering
	\includegraphics[scale=0.2]{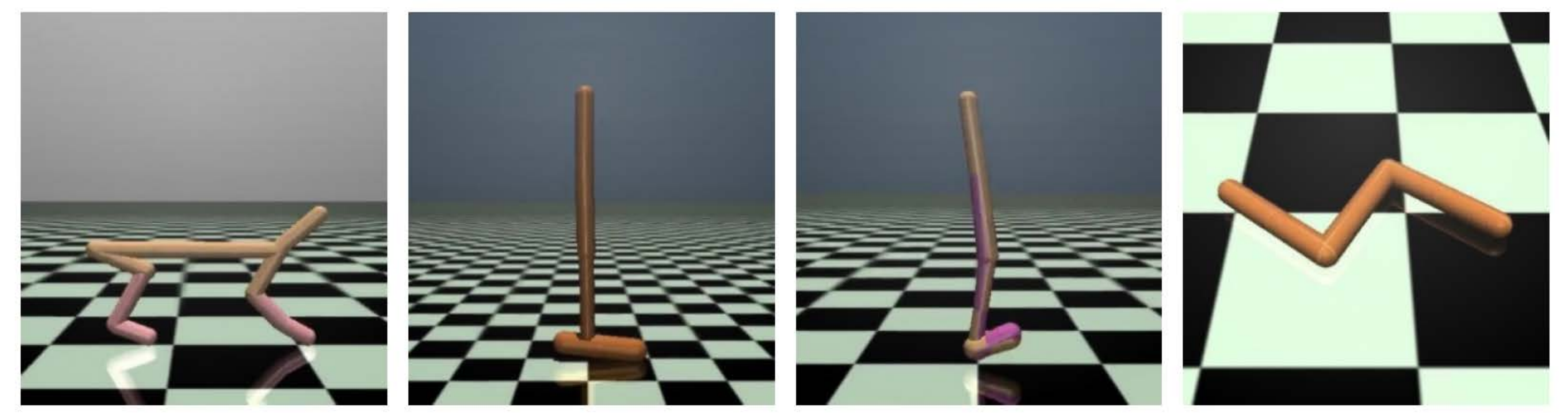}       
	\caption{Halfcheetah-v2, Hopper-v2, Walker2d-v2 and Swimmer-v2 tasks environment}
	\label{fig3}	
\end{figure}

MuJoCo \cite{Todorov14} stands for Multi-Joint dynamics with Contact. It is being developed by Emo Todorov for Roboti LLC, and has now been adopted by a wide community of researchers and developers in simulation tasks \cite{HaarnojaZAL18,WuMGLB17,SchulmanWDRK17,CiosekVLH19,LillicrapHPHETS15,MnihBMGLHSK16,SchulmanMLJA15,PintoDSG17}.
In experimental tasks of MuJoCo engine, according to the computer conditions, training time and task difficulty, we select four tasks with moderate difficulty which are able to show the experimental results from multiple aspects. The four simulation tasks are Halfcheetah-v2, Hopper-v2, Walker2d-v2 and Swimmer-v2, which are shown in Figure \ref{fig3}.

The main goal of training agents is to train an appropriate action selection policy according to state space data of each frame in environment, and continuously control the action of each joint in each frame, in order to complete fast and stable forward movement. After each frame's action, environment feedbacks output data, such as the observation of current state, whether the episode gets to terminal and the current frame's environment reward. In the process of movement, reward is composed of maintaining stability and moving forward. In some tasks, such as Walker2d and Hopper, if agent falls dowm, episode will end immediately, otherwise, it ends automatically every 1000 frames. Therefore, an agent needs to balance conservative policy of maintaining stability and receiving a small amount of reward continuously with aggressive policy of controlling joints that easily leads to early end but may receive more reward.

The Tensorflow-GPU framework is used to complete the efficient construction and training of neural networks. The basic architecture of the three algorithms in this paper is the same as PPO. Improved algorithms add curiosity module and uncertainty estimation module. Model construction of PPO algorithm is almost consistent with the introduction and configuration of OpenAI. The detailed configuration is as follows:

\begin{itemize}
	\item \textbf{Policy network.} Policy network completes the mapping from state space $S$  to action space $A$  through the three-layer fully connected neural network. The dimensions of the input layer and output layer are the same as state space and action space respectively. The dimensions of both the two hidden full connection layers are 64. Activation function is tanh. Gaussian action noise is sampled and added to the output layer to complete the action selection. Loss function based on action selection probability is shown in formula (\ref{8}). Learning rate is 0.0003.
	\item \textbf{Value function network.} Value function network completes the mapping of state space $S$ to state value $V$. Output layer is the one-dimensional cumulative reward estimation, and the dimensions of both the two hidden fully connected layers are 64. Activation function is tanh. Mean square error based on the real reward is used as the loss function to conduct the supervised learning optimization with a learning rate of 0.001.
	\item \textbf{Curiosity network.} Curiosity network takes state $s_{t}$ and action $a$ as input and it outputs the state $s_{t + 1}$ of the next frame through a 32-dimensional layer. Activation function is tanh. MSE based on the real observation state is used as the loss function to optimize the supervised learning with a learning rate of 0.001.
	\item \textbf{Uncertainty estimation network.} Uncertainty estimation network for two states $s_{t}$ and $s_{t + n}$ as input, with the output for the step estimations of the number $\hat{n}$ , and the activation function is tanh. By using mean square error, supervised learning optimization with learning rate of 0.001 is carried out.
	\item \textbf{Other Hyperparameters.} Optimization times of each training episode is 80; discount factor is 0.99; clipping ratio is 0.2.
\end{itemize}

In certain reinforcement learning experiment, corresponding adjustments according to the environment influence training effect. Some optimizations are not appropriate for all tasks, which is one of the reasons for reinforcement learning training is difficult to reproduce. Optimizations unrelated to algorithms in previously published literature may affect training results \cite{Gao0L18,abs-1906-10306,YeLSSZWYYWGCYZS20}. Such as data normalization, adding constraints, adjusting initialization settings, etc.

In order to ensure the accuracy and persuasion of the experiment, we would optimize the algorithm according to the experimental environment and process. So that our PPO algorithm training effect is better than the results when it was proposed \cite{SchulmanWDRK17}. The specific treatment is as follows:
\begin{itemize}
	\item \textbf{Normalize advantage value.} we use value function network to calculate advantage value $A$, which is used to evaluate the advantage of the selected action compared with other actions in the same state. Advantage value is also affected by the accuracy of value function network and real reward. Normalization processing can effectively optimize actions of the same batch in each episode of training. When reward value in the task varies greatly, normalization strategy can make whole training process more stable, but may increase training bias caused by nonuniform sampling.
	\item \textbf{Data processing.} Monte-Carlo method is used for data acquisition and error calculation. In this method, the reward of each frame affects cumulative reward value of the previous hundred of frames, which also enables the algorithm to take into account the potential influence of current action selection in subsequent multiple steps. However, if episode shuts down after current frame is completed, all subsequent reward will default to 0. A large bias is also introduced due to a fixed time being reached. This error not only affects previous frames but is difficult to eliminate. In a fixed timestep tasks, we set the last frame's reward as value function network output. The result improvement compared with the original setting of 0.
	\item \textbf{KL divergence constraint.} Although the negative effects are alleviated by ratio clipping of new and old policies, the problem of inappropriate step size caused by nonuniform sampling is revealed in the advantage value normalization. Therefore, KL divergence constraint is carried out. If average KL divergence is greater than preset constraint value in current episode of training, this training episode will be terminated. The preset value in experiment is 0.015. KL divergence constraint can make training process more stable.
\end{itemize}

\subsection{Gaussian noise exploration effect experiment} 
\begin{figure}[ht]
	\centering
	\includegraphics[scale=0.27]{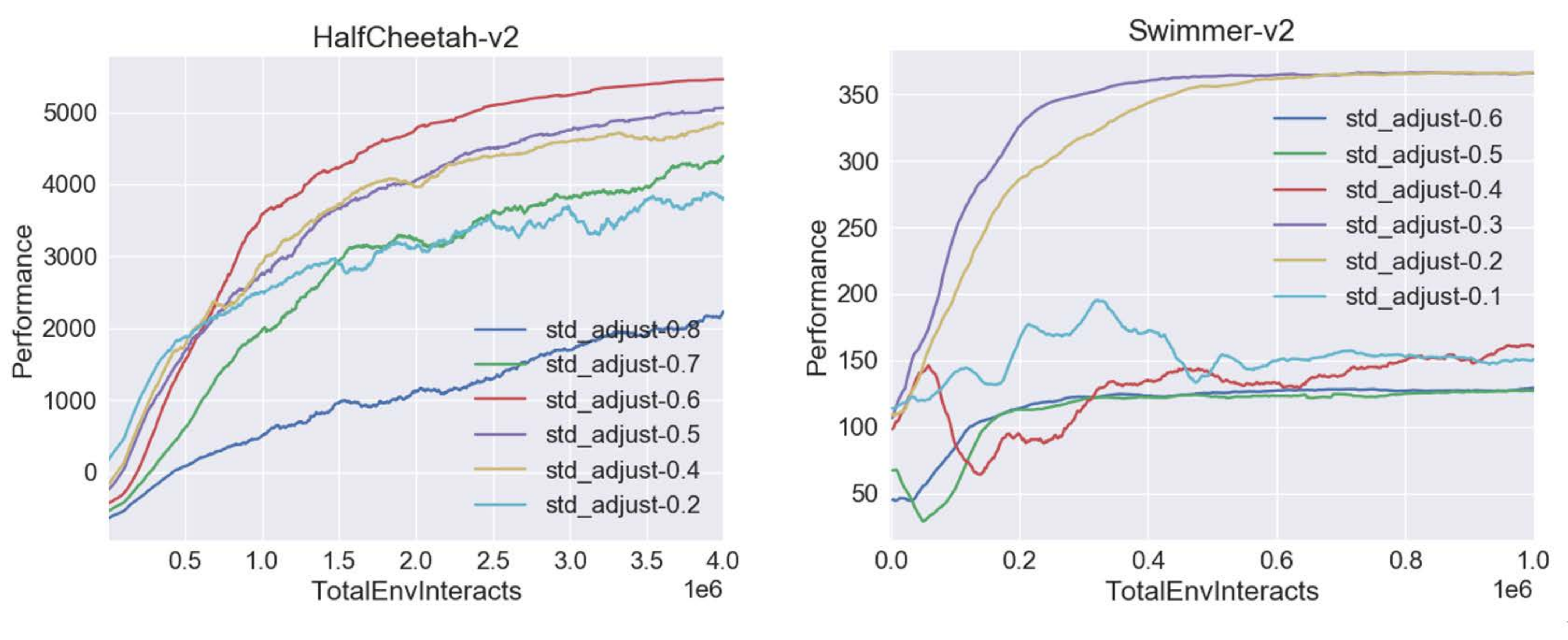}       
	\caption{Comparison of performance by various Gaussian exploration initial values}
	\label{fig4}	
\end{figure}

Gaussian noise exploration, as a basic exploration method of policy gradient algorithm, has certain exploration effect on different continuous action tasks. The mechanism of dynamically adjusting Gaussian noise based on cumulative reward can have a stronger exploration scope in early stage of training and a higher rewarding efficiency in late stage. Here, we compare and verify the setting of multiple initial Gaussian noise $\sigma$ of Halfcheetah-v2 and Swimmer-v2 tasks. The experimental results are shown in Figure \ref{fig4}.

Figure \ref{fig4} shows the performance of experiments on two different tasks with different initial Gaussian noise  $\sigma$  values. Six different settings are selected for each task, which can be seen from training trend and reward of different settings:
\begin{enumerate}[fullwidth,itemindent=2em,label=\arabic*)]
	\item In early stage of training, the smaller exploration scope (i.e., Gaussian standard deviation) has a better performance (performance in the first 13\% interacts for HalfCheetah and 2\% for Swimmer), but at last, premature convergence leads to the suboptimal. Too large Gaussian standard deviation leads to policy degradation, which cannot be optimized to the optimal or takes too much time. Therefore, the setting of Gaussian noise exploration should be adjusted from two aspects of training speed and the final performance.
	\item  Considering factors of speed and performance, it can be clearly found that in HalfCheetah, the best performance will be achieved if initial value of $\sigma$ is set near 0.6. In both 0.5 and 0.7, the performance is significantly damaged. Relatively in Swimmer, a range of 0.2 to 0.3 would work equally well and converge to optimal value, and with other settings the policy would fall into a local optimum. Therefore, it is necessary to adjust the action exploration settings according to specific tasks, as the performance is affected by the exploration ability which is related to difficulty and type of tasks.
\end{enumerate}

\begin{figure}[ht]
	\centering
	\includegraphics[scale=0.27]{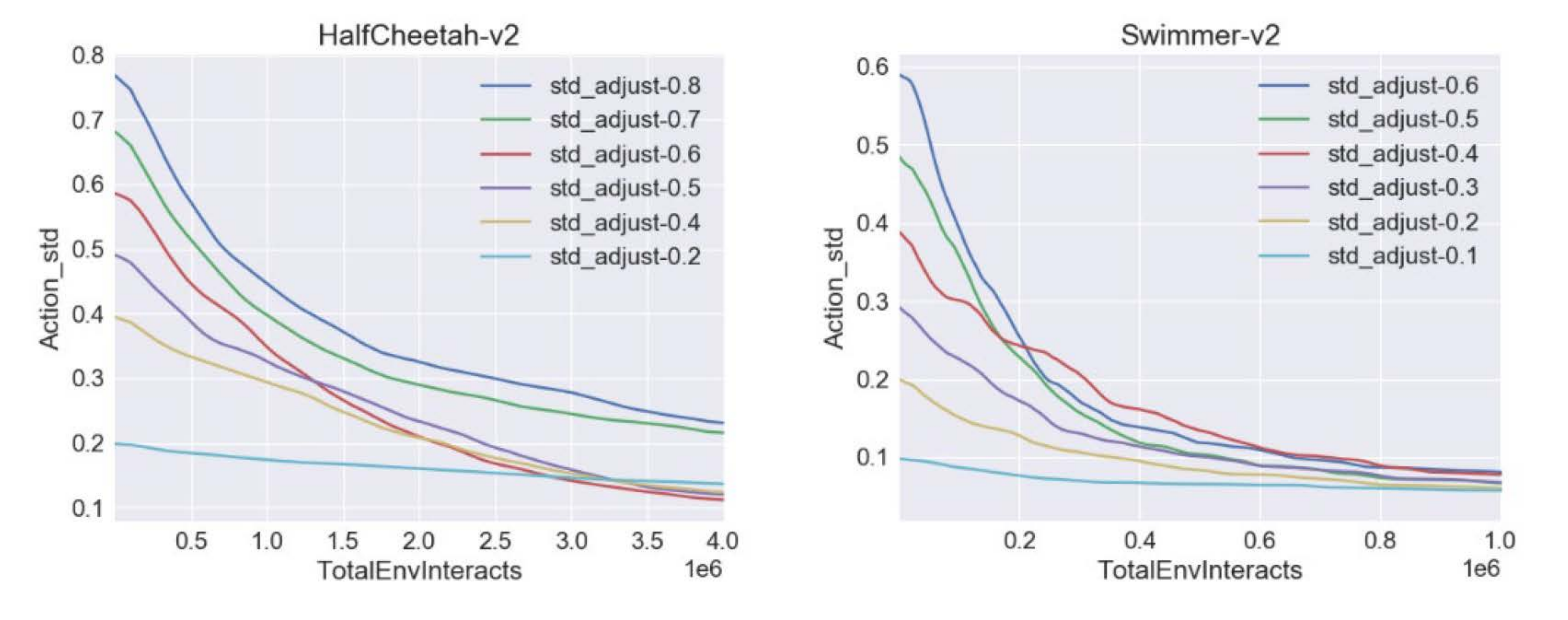}       
	\caption{Changes in the exploration scope of various Gaussian exploration initial settings}
	\label{fig5}	
\end{figure}
Figure \ref{fig5} shows the changes in the scope of Gaussian action exploration in two tasks under different initial values. The ordinate represents the change of $\sigma$ value in Gaussian noise.

As can be seen from the experimental results, if policy has converged (the initial value of $\sigma$ in HalfCheetah is 0.2$ \sim $0.6, because the cumulative reward of 0.7 and 0.8 is still on the rise in the later stage of training. And the initial value of $\sigma$  in Swimmer is 0.1$ \sim $0.6), exploration scope converge almost to the same range (convergency value of $\sigma$  in HalfCheetah is 0.16 and in Swimmer is 0.06). Therefore, the Gaussian noise of different experimental settings mainly affects performance by exploration ability and learning progress. The performance will not affect by the different noise scope after the training convergence.

\subsection{Exploration enhancement algorithm experiment} 	

We demonstrate two directional exploration mechanisms. One is curiosity driven based on prediction error as intrinsic incentive, which is called as ICM-PPO algorithm. The other is uncertainty measurement based on step size as intrinsic incentive, which is referred to as IEM-PPO algorithm. We will carry out a variety of experimental comparison to illustrate effectiveness and advantages.

\subsubsection{Exploration enhancement mechanism on training effect} 

\begin{figure}[ht]
	\centering
	\includegraphics[scale=0.27]{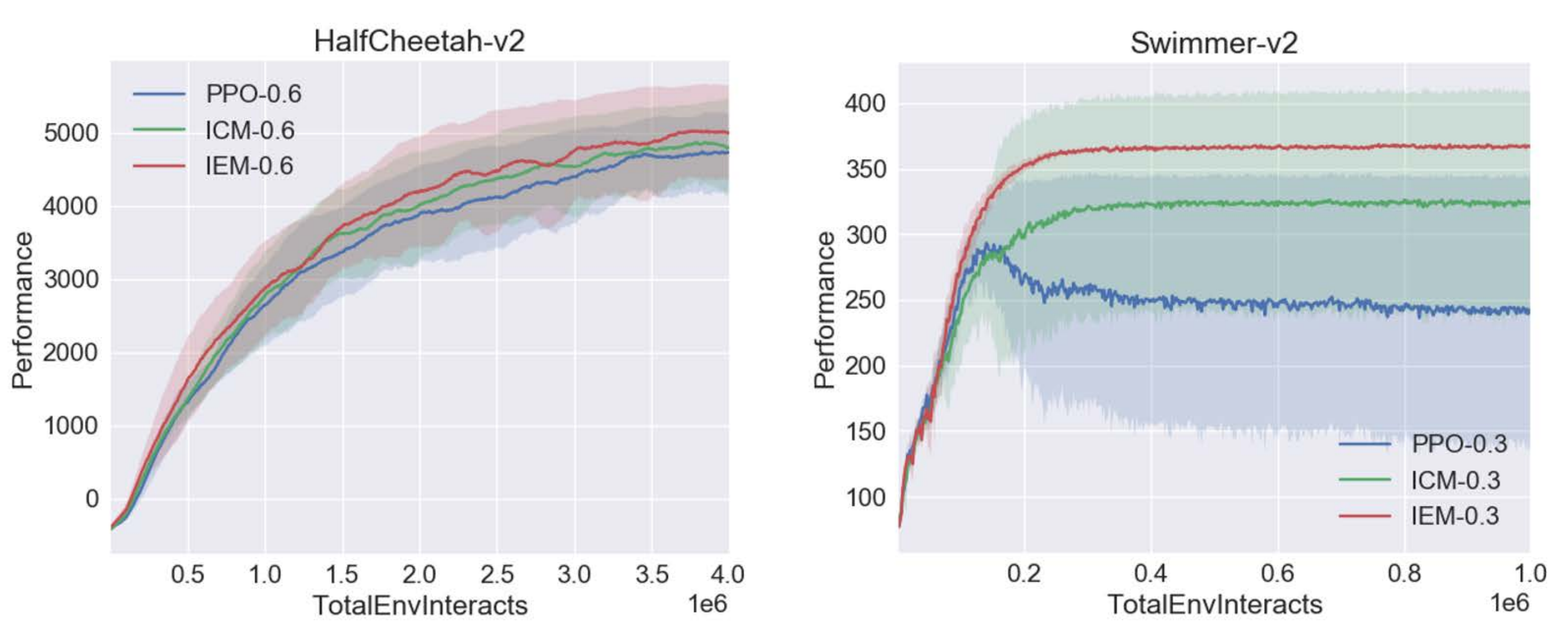}       
	\caption{Comparison of performance of PPO, ICM-PPO and IEM-PPO algorithms in two environments}
	\label{fig6}	
\end{figure}

Figure \ref{fig6} shows the performance of two exploration mechanisms when combined with PPO algorithm, in which does not adjust other parameters except for the intrinsic incentive on the reward value. Therefore, under the optimal hyperparameters setting of PPO algorithm, the effectiveness of the exploration mechanisms can be tested through experimental comparison.	

For each of three algorithms, the same experiment is conducted five times, and the contingency of results are eliminated by taking mean value. Solid line shows the change of average cumulative reward value corresponding to the number of interactions with environment, and shaded part shows variance fluctuation of cumulative reward. It is clear that the performance of IEM-PPO algorithm is higher than other two algorithms (converging near 5050, 4880, 4800 in HalfCheetah and 367, 320, 250 in Swimmer), and has better stability in some tasks (data in Swimmer with a small shadow range).

Thus, it can be concluded that IEM-PPO algorithm is outperforming to both ICM-PPO and PPO algorithms in terms of training efficiency (performance in whole training process) and final performance at the end of training. It can also be seen that the exploration mechanisms such as IEM or ICM can enhance exploration efficiency without hindering the exploitation effect, so as to obtain a faster training speed and find a better solution.

\subsubsection{Interaction between exploration mechanisms} 
\begin{figure}[ht]
	\centering
	\includegraphics[scale=0.35]{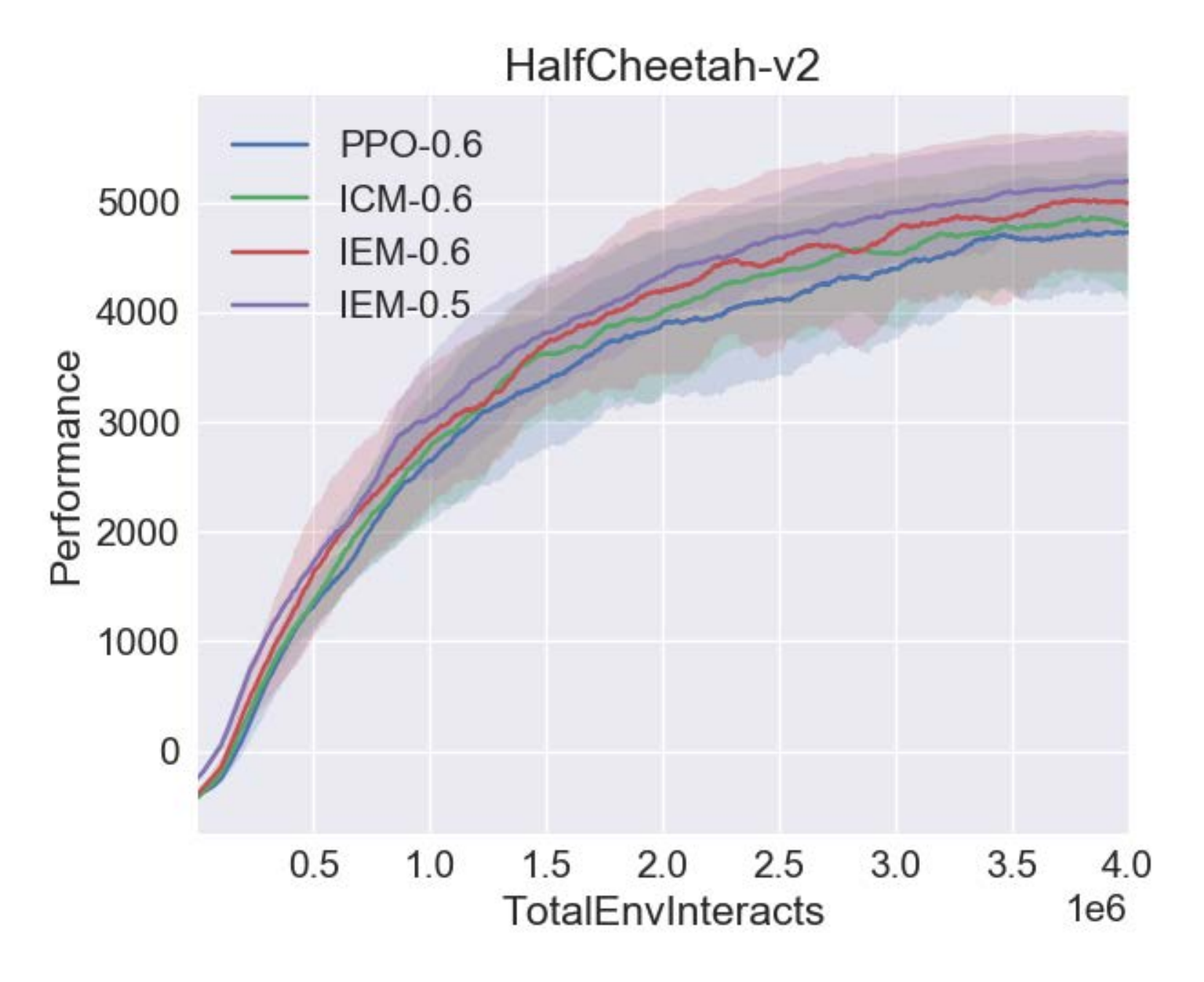}       
	\caption{Performance of IEM-PPO algorithm for exploration scope tuning}
	\label{fig7}	
\end{figure}

In the experiment with the same setting as the PPO, the effect of exploration enhancement mechanism is demonstrated. However, theoretically speaking, Gaussian noise can explore around action, but in exploration enhancement mechanism, the direction of exploration is changed by changing the reward value of training process. The two exploration mechanisms interact with each other, exploration enhancement mechanism can ensure Gaussian noise in a smaller and more accurate scope. In previous Halfcheetah-v2 experiment, it has been concluded that performance of this task is sensitive to Gaussian noise setting. Therefore, after adding IEM module here, adjusting Gaussian noise setting will have a better effect, as shown in Figure \ref{fig7}.

Figure \ref{fig7} shows IEM-PPO algorithm with adjusting the initial value of Gaussian noise $\sigma$ to 0.5. It can be seen that after adjusting the initial value, a higher average cumulative reward (about 5200) is obtained, and a higher performance is achieved in the whole training process. Therefore, exploration enhancement mechanism interacts with Gaussian noise, which will have better training effect after appropriate tuning.

Although the hyperparameter setting can be tuned to achieve better performance, the introduction of enhanced exploration model increases complexity of hyperparameter tuning. In this experiment, it only demonstrates that there are settings to obtain better performance, we does not search optimal hyperparameter in whole hyperparameter range.

\subsubsection{Stability of exploration enhancement mechanism} 
We demonstrate the stability of our algorithm in this subsection. We employ Halfcheetah-v2 task for testing because the hyperparameters effects performance obviously \cite{PintoDSG17}. After comparing performance of various settings, the results are shown in Table \ref{table:table3}.

Table \ref{table:table3} lists the changes of IEM-PPO algorithm and PPO algorithm under different size exploration scope settings. We repeat the experiment three times for each setting, and the performance is averaged after every 100 episodes.  IEM-PPO algorithm has higher performance in all hyperparameter settings and has the best convergency value 5252.09. In addition, IEM-PPO algorithm has 5.17\%, 23.07\%, 18.66\% and 5.19\% improvements respectively with the compare of PPO algorithm, and the influence by parameters is slightly lower than the PPO algorithm (IEM-PPO performances on the initial $ \sigma $ value of 0.6$\sim$0.4 are all well). Therefore, the algorithm with exploration enhancement mechanism is more stable.

\begin{table}[H]
	\setstretch{1.2}
	\begin{center}
		\caption{Comparison of IEM-PPO and PPO under various settings in Halfcheetah-v2}
		\label{table:table3}
		\begin{tabular}{cccccc}
			\toprule
			 & adjust-0.6 & adjust-0.5 & adjust-0.4 & adjust-0.2 \\ \midrule
			PPO         & 4824.30    & 4267.53    & 4242.58    & 4084.10    \\
			IEM-PPO     & \textbf{5073.83}    & \textbf{5252.09}   & \textbf{5034.14}    & \textbf{4296.04}   \\
	 \bottomrule
		\end{tabular}
	\end{center}
\end{table}

\subsubsection{Experimental effects in more tasks} 

Through the experiments on Halfcheetach-v2 and Swimmer-v2, we have explained the influence of exploration enhancement mechanism on the training performance and stability. Here, we'll expand into more environments to see how well the IEM-PPO algorithm fits into various tasks. Because MuJoCo's various simulation tasks are different in task objectives, environmental reward calculation, episode terminal, friction and gravity mechanism, etc. There are diversified optimization policies in different tasks, which can test the feasibility of the algorithm in various aspects. The experimental results are shown in Table \ref{table:table4}.

\begin{table}[H]
	\setstretch{1.2}
	\begin{center}
		\caption{Comparison of experimental performance of three algorithms in more tasks}
		\label{table:table4}
		\begin{tabular}{cccccc}
		\toprule
		&                &HalfCheetah&Swimmer& Hopper & Walker2d  \\ \midrule
		& Interacts & 4000*1000   & 2000*500 & 2000*500 & 4000*1000 \\ \hline
		\multirow{2}{*}{PPO}     & Reward         & 4824.30     & 242.33   & 2059.52  & 2760.64   \\
		& Variance       & 544.97      & 4.55     & 882.98   & 1202.87   \\  \hline
		\multirow{2}{*}{ICM-PPO} & Reward         & 4834.10     & 324.16   & 2018.37  & 2801.26   \\
		& Variance       & 570.57      & 2.58     & 827.87   & 1213.52   \\ \hline
		\multirow{2}{*}{IEM-PPO} & Reward         & \textbf{5073.83}     & \textbf{367.74}   & \textbf{2158.52}  & \textbf{2971.12}  \\
		& Variance       & \textbf{359.92}     & \textbf{1.89}     & \textbf{770.43}   & \textbf{1107.87}  \\ \bottomrule
		\end{tabular}
	\end{center}
\end{table}

Table \ref{table:table4} shows the average performance and variance of various algorithms in tasks. Among them, the number of interactions with environment in task is different (4000*1000 and 2000*500), which is mainly determined by the difficulty of the environment. Moreover, the number of training times in each task has enabled the policy to converge. All data in table is experiment for three times and get real average reward value in final 100 episodes of policy. It can be seen that IEM-PPO algorithm is outperforming to other algorithms. Besides having higher performance, it also has more stable polices and certain generality for various tasks.

\subsubsection{Time complexity of exploreation enhancement mechanism} 

\begin{figure}[ht]
	\centering
	\includegraphics[scale=0.35]{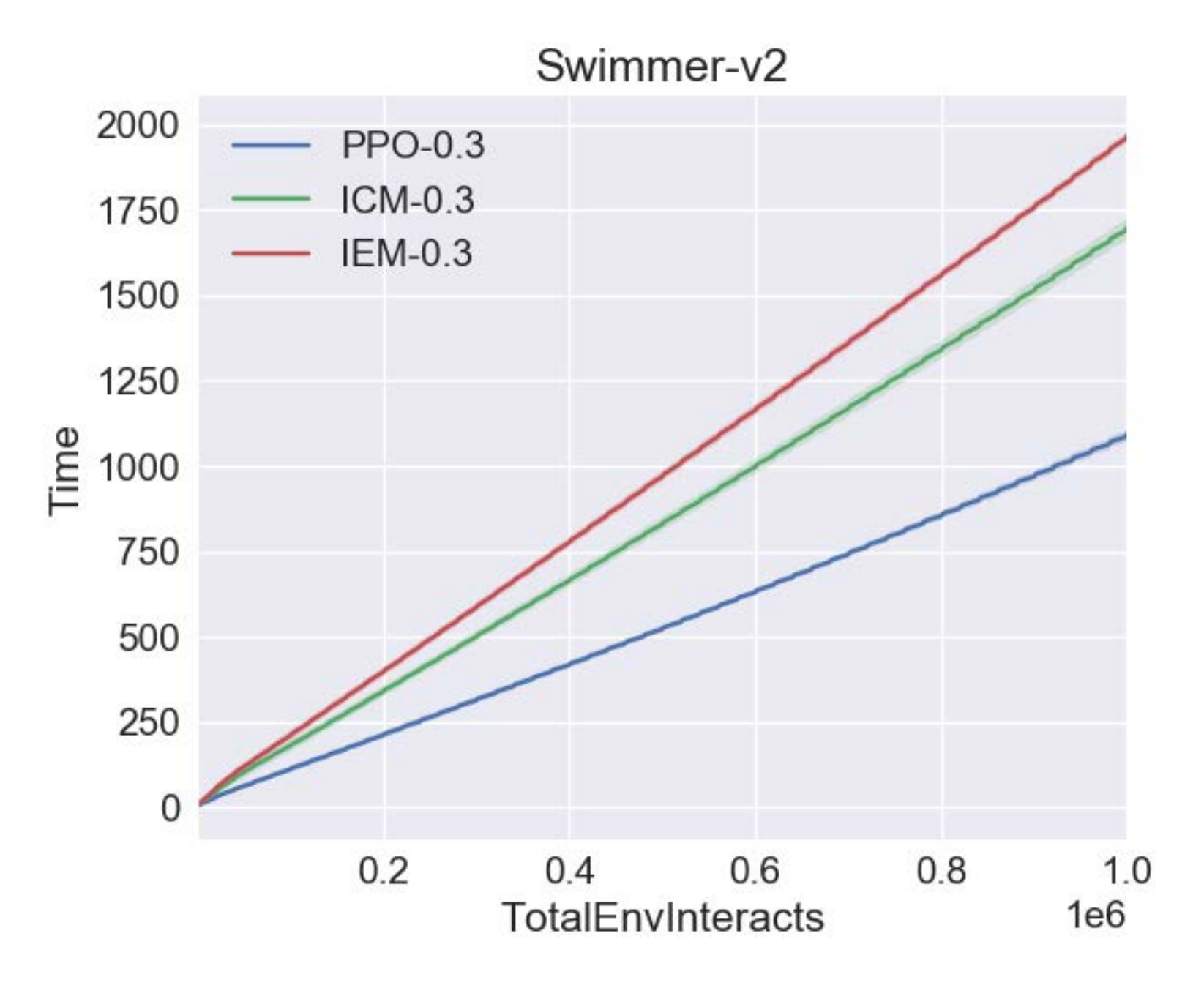}       
	\caption{Comparison results of the algorithms in training duration}
	\label{fig8}	
\end{figure}

The exploration enhancement mechanism is designed to identify potential relationships and trends from the limited available data. In the training process, the scale of the neural network is increased and the internal incentive needs to be calculated in each frame, so the training time is longer. Comparison of training time is shown in Figure \ref{fig8}.

The total time spent training an agent using different algorithms in Swimmer-v2 environment is shown in Figure \ref{fig8}. It can be seen that in training process, the IEM-PPO and ICM-PPO algorithms need longer computing time, but both of them are proportional based on environmental interactions. Therefore, although enhancement mechanisms has increased the computational burden of the training process, it has the same complexity with PPO algorithm, and computational time is still within the acceptable range. Note that training time is not only affected by algorithm framework, but also affected by environment complexity, hardware computing speed and code implementation efficiency. 

\section{Conclusion and Future Work} 

In this paper, we propose a new algorithm based on exploration enhancement mechanism, and demonstrate algorithm effects in continuous action tasks.

Firstly, we theoretically analyze the function and defect of Gaussian exploration mechanism in training process of PPO algorithm. From practical view, experimental verification of various exploration scope settings shows that appropriate and efficient exploration settings should be adopted to ensure performance for different environments.

Then, from the perspective of improving efficiency of algorithm exploration, ICM-PPO algorithm based on curiosity driven exploration is implemented and IEM-PPO algorithm based on uncertainty estimation is proposed. Based on uncertainty estimation theory, IEM-PPO algorithm uses collected environmental state data to construct neural network to complete the uncertainty estimation function, using uncertainty estimation as internal incentive to carry out positive incentive for the action exploration. We ensures efficient exploration under the condition of considering training time.

In experiment, we use PPO algorithm and ICM-PPO algorithm for comparison with IEM-PPO algorithm on Mujoco physical simulation environment. Considering the exploration ability, IEM-PPO algorithm improves training speed and final training performance. In terms of stability, exploration enhancement mechanism is applicable in most parameter settings and has stability. In training time, the new neural network increases the scale of the model and the number of parameters in the model which requires longer training time but not increases too much computing burden.

We can conclude that although IEM-PPO algorithm requires longer training time, it has excellent training efficiency and performance, and has stability and robustness.

Future work can be explored from the following three aspects:
\begin{enumerate}[fullwidth,itemindent=2em,label=\arabic*)]
	\item In discrete action tasks, if the optimal action distribution is bimodal or multimodal, policy gradient algorithm will have better effect. However, the Gaussian distribution in continuous action tasks can only be a single peak, which cannot give full play to the potential advantages of policy gradient algorithm.
	\item The dimension of internal incentive is not consistent with extrinsic reward. For reinforcement learning task, variation range of reward corresponding to the good and bad policies also affects the effect of the exploration enhancement mechanism. Therefore, a general strategy is needed to solve hyperparameter optimization problem \cite{ZhaLZH19,PintoDSG17}.
	\item It may be better to use the characteristics of multi-agent learning for further exploration \cite{SongRSE18,SalimansHCS17,WangWHZG16}, but the current multi-agent learning is weaker than single agent under the same amount of training, so it is necessary to balance the optimization objectives of multi-agents in an appropriate way.
\end{enumerate}

\section*{Acknowledgement} 

This work was supported by the National Key R\&D Program of China under Grant No. 2017YFB1003103; the National Natural Science Foundation of China under Grant Nos. 61300049, 61763003; and the Natural Science Research Foundation of Jilin Province of China under Grant Nos. 20180101053JC, 20190201193JC.


\bibliography{mybibfile}

\end{document}